%% file: main_paper.tex
\documentclass[runningheads]{llncs}
\usepackage{url}
\usepackage{comment}
\usepackage{amssymb}
\usepackage{booktabs} 
\usepackage{enumitem}
 \usepackage{array,multirow,graphicx}
 \usepackage{wrapfig,lipsum}
 \usepackage{float}
\usepackage{pgfplots}
\usepackage{csquotes}
\usepackage{makecell}

\usepackage{tikz}

\usepackage{xspace}
\newcommand{\PModel}{\textit{FFM}\xspace}
\newcommand{\HighConcern}{HiPC\xspace}
\newcommand{\MediumConcern}{MePC\xspace}
\newcommand{\LowConcern}{LoPC\xspace}

\usepackage[symbol]{footmisc}

\usepackage{algorithm}
\usepackage{algcompatible}
\usepackage{amsmath}
\usepackage{amsfonts}
\usepackage{subfig}
\newcounter{Figcount}
\newcounter{tempFigure}

\usepackage[colorinlistoftodos]{todonotes}

\usepackage{comment}
\newcommand{\Lili}[1]{\textcolor{red}{[Lili: #1]}} 
\newcommand{\sharma}[1]{\noindent\textcolor{red}{\today:~}\textcolor{blue}{ \textbf{~Sharma: #1} \\}} 

\begin{document}

\title{Flexible and Efficient Multi-Feature Data Analysis With both Content and Structure}
\title{Hello World: Multiplex Network Data Analysis with the Fusion of Content and Structure}
\title{Generic Multilayer Network Data Analysis with the Fusion of Content and Structure}


\institute{}
\titlerunning{Generic Multilayer Network Data Analysis [...]}
 \author{Xuan-Son Vu$^1$, Abhishek Santra$^2$, Sharma Chakravarthy$^3$, Lili Jiang$^1$}
 \authorrunning{Xuan-Son Vu et al.}
 \institute{$^1$Department of Computing Science, Ume\r{a} University, Sweden;\\
 $^{2,3}$Information Technology Laboratory, University of Texas at Arlington, USA;\\
 \email{$^1$\{sonvx, lili.jiang\}@cs.umu.se,
 	$^2$abhishek.santra@mavs.uta.edu,$^3$sharma@cse.uta.edu
 }
 }

\maketitle


 
%


\begin{abstract}
\vspace{-25pt}
Multi-feature data analysis (e.g., on Facebook, LinkedIn) is challenging especially if one wants to do it efficiently and retain the flexibility by choosing features of interest for analysis. Features (e.g., age, gender, relationship, political view etc.) can be explicitly given from datasets, but also can be derived from content (e.g., political view based on Facebook posts).  Analysis from multiple perspectives is needed to understand the datasets (or subsets of it) and to infer meaningful knowledge. For example, the influence of age, location, and marital status on political views may need to be inferred separately (or in combination).

In this paper, we adapt multilayer network (MLN) analysis, a non-traditional approach, to model the Facebook datasets, integrate content analysis, and conduct analysis, which is driven by a list of desired application based queries. 
Our experimental analysis shows the flexibility and efficiency of the proposed approach when modeling and analyzing datasets with multiple features.
 
\end{abstract}


%



\input{ccs-body} 

\bibliographystyle{splncs04}
\bibliography{xuansonResearch.bib,santraResearch.bib}

\end{document}

%% file: ccs-body.tex
\input{ses1_intro_sharma}


\section{Related Work} \label{sec:related-work}
\input{ses2_relatedwork_abh}
\input{ses2_relatedwork_svx}

\input{ses3_problemstatement}


\input{ses4_modeling_and_approach_svx}
\input{ses4_modeling_and_approach_abh}

\input{ses5_content_analysis_svx}

\input{ses6_analysis}
\input{ses7_conclusions}

%% file: ses1_intro_sharma.tex
\section{Introduction}

Analysis of complex datasets containing multiple heterogeneous features such as numeric, categorical, text-based features etc. has become relevant.  Social networks (e.g., Facebook, Twitter etc.) fall in the category, and a framework that seamlessly integrate multiple features is desired.  In this paper, our goal is to adapt an approach that allows us to efficiently and flexibly analyze social network data using explicit (known or given) as well as implicit (derived or extracted) features of the datasets. It is imperative that these datasets be analyzable in a flexible manner as different features have different impacts and importance on the information that can be inferred. For example, for advertising in social networks, influential communities are sought (based on age, gender, friends, interests, political views etc.). For quick propagation of information centrality nodes may be useful. The approach taken in this paper is generalized although illustrated on a specific real-world, large data collection. 

Several approaches are available for analyzing datasets, each with its own advantages and limitations. Database Management systems (DBMSs) have been around for a while and they are good at answering \textit{exact queries}. They are good for analysis that needs exact answers from the dataset. Statistical, mining, and other approaches have been around, also for quite a while, and they are good for understanding aggregate characteristics -- an example being one that is carried out on census datasets to extract inferences and trends over time periods. For datasets that have relationships as part of them, graph modeling and mining became useful and a number of analysis techniques have been developed (e.g., clustering, communities, hubs, triangles,...). As the datasets have become more complex in terms of the number of features, modeling them as a single graph makes the graph complex -- with multiple edges between nodes corresponding to different relationships, multiple node \& edge labels, etc. Modeling becomes complex and applying traditional algorithms for flexible analysis is either impossible or more complex and inefficient. We term this as the \textit{traditional approach.}

The approach used in this paper, termed \textbf{M}ulti\textbf{L}ayer \textbf{N}etwork (MLN) analysis with decoupling, is in its early stages and being researched actively. The MLN approach
does not change the analysis, \textit{except how} datasets are modeled and analyzed. It has been receiving a lot of attentions in the last decade due to its advantages: i) \textit{allows modeling of a complex dataset using a set of user-definable simple, single graphs termed layers) }, ii) \textit{allows the same analysis as the traditional approach on this model without loss of accuracy}, iii) is amenable to parallelism (for scalability) and has been shown to be better in storage requirements and efficiency. There are other advantages as well~\cite{ICCS/SantraBC17,MultiLayerSurveyKivelaABGMP13}.


In this paper, we plan to 
showcase the above advantages of MLN approach. This paper is the first one, to the best of our knowledge, to apply this approach for the analysis of \textit{one of the largest/densest real-world social network data collection}, although it has been used in several experimental studies on smaller/sparser datasets \cite{cardillo2013emergence,MinSubMulLayer2012}.

The contributions of this paper include:
\textbf{(1)} using a novel, emerging MLN approach for flexible analysis of a large complex real-world dataset, \textbf{(2)} establishing its modeling benefits, flexibility of analysis, and efficiency of computation, \textbf{(3)} integrating content analysis seamlessly with structural network analysis, and \textbf{(4)} extensive analysis and result validation for the social network work datasets.

The remainder of the paper is organized as follows. Section~\ref{sec:related-work} discusses related work. Section~\ref{sec:problem-statement} states the general problem and proposed approach. Section~\ref{sec:modeling-and-computation} elaborates modeling and computation aspects of our approach. Section~\ref{sec:content-analysis} details the use of content analysis to integrate into multilayer network approach. Section~\ref{sec:analysis-results} showcases analytical results of  queries. Section~\ref{sec:efficiency-evaluation} discusses computational advantage of the adapted approach. Section~\ref{sec:conclusions} concludes the paper.

%% file: ses2_relatedwork_abh.tex
We provide an overview of some related work including multilayer networks, community detection, and content-based analysis.
\\
\noindent\textbf{Multilayer Networks or Multiplexes:} 
Significant amount of work has been done in the area of multilayer networks~\cite{MultiLayerSurveyKivelaABGMP13,Boccaletti20141}.  They have been categorized into homogeneous, heterogeneous, and hybrid multiplexes depending on whether the layers have same, different, or a combination of entity sets, respectively \cite{BDA/SantraB17}. In this paper, we focus on homogeneous multiplexes where the entity set (i.e., people in our case) is fixed, but the features are numerous giving rise to a layer for each feature. Considerable research has been done to handle varying interactions among the same set of entities such as co-authorship in different conferences \cite{MinSubMulLayer2012} and city connectivity based on different airlines \cite{cardillo2013emergence}. 
In order to understand the effect of multiple features using composition of 
multiplex layers, a principled approach \cite{ICCS/SantraBC17,ICDMW/SantraBC17} has been proposed to arbitrarily combine features (or layers) and then analyze them without having to construct combined layers. The composition approach leads to efficient (both computation and storage) and flexible analysis of combinations of features using Boolean operations.
\\
\noindent\textbf{Community Detection} is a well-studied problem in monoplex (single network) analysis. It involves identifying groups of nodes that are more connected to each other than to other nodes in the network. 
Some work has investigated community detection methods for multilayer networks \cite{CommSurveyKimL15,Mucha10}. 
But there is hardly any work for determining the communities for different multiplex layer compositions, until recently where a novel approach was proposed to infer communities of a combined network from communities of individual layers using Boolean compositions \cite{ICCS/SantraBC17}. We posit in this paper that this approach is appropriate for analyzing multi-feature datasets flexibly and efficiently.
\\

%% file: ses2_relatedwork_svx.tex
\noindent\textbf{Content Analysis} is a research method for studying documents and communication artifacts, which might be texts of various formats, pictures, audio or video \cite{Pashakhanlou2017}.  One of the key advantages of using content analysis to analyze social phenomena is its non-invasive nature, in contrast to simulating social experiences or collecting survey answers. In this paper, we apply content analysis by text mining techniques to integrate content features into multilayer structural analysis. Motivated by the fact that, we have personality data in the given Facebook data collection and the previously preliminary studies \cite{Sumner:2011,Vu:2018b} observed a significant correlation between personality and privacy-concern of Facebook users, we improve \cite{Vu:2018b}'s deep neural network model in this paper to detect Facebook users' privacy-concern based on their status updates and classify privacy-concern into three different levels (classes): \textit{high, medium}, and \textit{low}. These content-derived classes are treated as content features and modeled as layers to integrate into the multilayer structural analysis.

%% file: ses3_problemstatement.tex
\section{Datasets, Analytical Queries, and Problem Statement}
\label{sec:problem-statement}
\subsection{Datasets}
\label{subsec:dataset}
For this paper, we chose the Facebook (FB) data collection to showcases the power and flexibility of the MLN approach,  since the data collection fulfill all the characteristics of a complex dataset in terms of a large number of features, content that could be analyzed and the requirement for analysis using combinations of features.

This data collection from \textit{myPersonality} project \cite{mypersonality:2015} is one of the largest and well-known real-world social network research data collection, where the volunteers took real psychometric tests and opted in to share data from their FB profile (period of 2007-2012).  The experimental data contains four datasets:  Demographic Info (D1), User's Political Views (D2),  Personality (D3), and FB Status Updates (D4).  We have a total of 260K individuals in the datasets predominantly from USA. Of those, about 2.6K have more common features as compared to others. Hence, as shown in Table~\ref{tbl:simple_statistics_dataset}, we have chosen the 2.6K subset for detailed analysis. 

\begin{wraptable}{r}{0.49\textwidth}\centering
\vspace{-20pt}
	\centering
	\caption{Statistics of four datasets} 
\vspace{-10pt}
		\begin{tabular}{m{4.7cm}|m{1.5cm}|m{4.1cm}}
			\hline
			\textbf{Datasets} & \multicolumn{2}{l}{ \#users} \\ \hline
			Demographic Info (D1) & \multicolumn{2}{l}{2,676} \\
			User's Political Views (D2)	& \multicolumn{2}{l}{2,695} \\
			Personality (D3) & \multicolumn{2}{l}{2,485} \\	
			Facebook Status Updates (D4) & \multicolumn{2}{l}{1,645} \\
			\hline 
		\end{tabular}
	
\vspace{-20pt}	\label{tbl:simple_statistics_dataset}
\end{wraptable}

\vspace{0.2cm}



We use four features from dataset D1 including age, gender, relationship status, and locale. One self-declared political-view feature from D2. D3 provides five features which are the five personality traits of the Five-Factor Model (\PModel) \cite{Mairesse:2007}. \PModel is considered the most influential and standard model for personality trait prediction in psychology over the last 50 years. Based on D3 together with D4, one more feature of people's privacy-concern is inferred. For better understanding in later sections, we especially introduce the well-known five personality traits \cite{Costa:2008}, which are defined as openness to experience, conscientiousness, extraversion, agreeableness, and neuroticism.    

\begin{enumerate} [label=]
\item {Openness (OPN) to experience}: intellectual, insightful vs. shallow

\item Conscientiousness (CON): self-disciplined, organized vs. careless
\item Extraversion (EXT): sociable, playful vs. aloof, shy
\item Agreeableness (AGR): friendly, cooperative vs. antagonistic, faultfinding
\item Neuroticism (NEU): insecure, anxious vs. calm, unemotional
\end{enumerate}

\begin{figure}[ht]
	\renewcommand{\figurename}{Figure}
	\centering
	\includegraphics[height=0.23\textheight, width=0.98\textwidth]{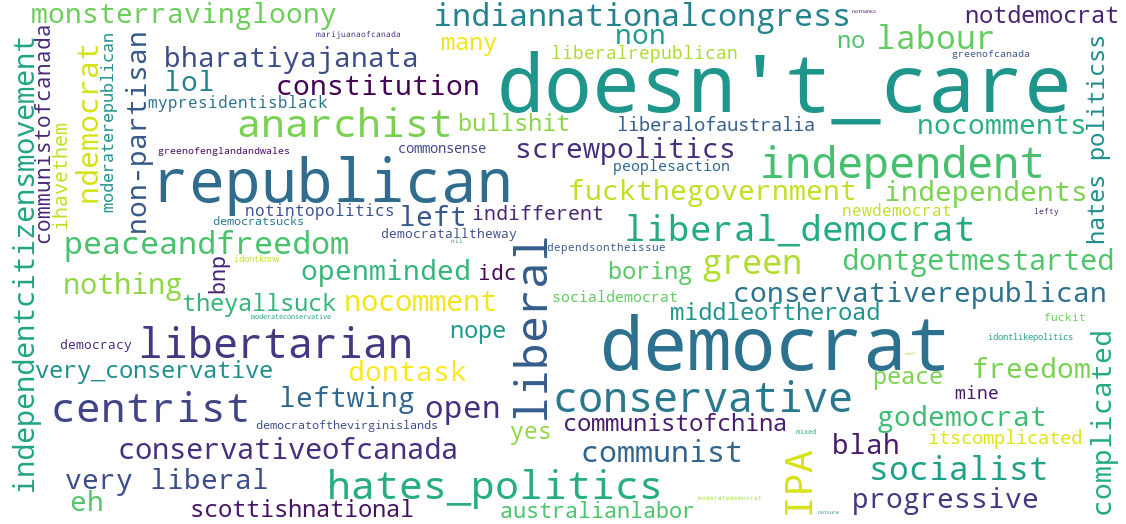}
	\caption{Word clouds of political views in the dataset in which bigger word represents a larger propotion.}
	\label{fig:political_views_cloudtag}
\end{figure}

\subsection{Analytical Queries} 
\label{subsec:query}
The focus of this paper is to demonstrate  modeling, flexible and efficient analysis of complex datasets to infer trends and establish correlations, and possible causality.  A few analytical queries that are meaningful for this dataset are shown below.

\begin{enumerate}[label=(Q\arabic*)]
\item \textbf{Dominant Political Views:} How the user-declared political view (e.g., democrat, doesn't care, republican) varies across age groups in the dataset? There are $100$ political views in Facebook dataset used in this paper, where ``doesn't care'' and ``democrat'' dominate others (see Figure~\ref{fig:political_views_cloudtag}).
 
\item \textbf{Relationship Status Correlation: } \begin{enumerate}
\item With respect to age groups, how does relationship status (e.g., single, in a relationship, and married) vary? 
\item How do the relationship statuses affect the personality traits of an individual? Does it differ based on gender?
\end{enumerate}
 
\item \textbf{Personality Trait Analysis:} 
\begin{enumerate}
\item How much of the population demonstrate contrasting personality traits (e.g., OPN and NEU)? 
\item How do the personality traits evolve with age? For example, which age group of people deals better with stress?
\end{enumerate}

\item \textbf{Privacy Concern Correlation:}  
\begin{enumerate}
\item How does the individuals' age correlate with their comfort level of sharing personal information on social media?
\item Do personality traits have a bearing on the level of privacy-concern?
\end{enumerate} 

\end{enumerate}

\subsection{Problem Statement}
\label{subsec:problem-statement}
Currently, the above analytical queries are done using a graph-based approach where \textit{a single graph needs to be created for each analysis based on the involved features}. Typically, nodes represent entities (i.e., people in our case) and edges represent relationship between nodes based on feature values (e.g., same age group, the same relationship status). These graphs are analyzed using graph metrics such as community, hubs or centrality nodes, and so on. This approach entails the creation of a customized graph for each query using the features involved which can lead to an exponential number of graphs in the worst case. For the above mentioned analytical queries (Q1-Q4), multiple graphs, the number of which depends on the number of features involved, need to be created, stored, and analyzed for each query. 


Using the MLN approach, \textit{for a given dataset with M features (whether explicit or derived), determine the layers based on the analyses requirements of the dataset and use composition for analysis using Boolean or other operations}. Thus, once multiplex layers are created as shown in this paper, any number of analyses can be performed without generating additional graphs/layers. Necessitated by the queries, this paper primarily uses \textit{AND} compositions.

%% file: ses4_modeling_and_approach_svx.tex
\section{Data Modeling and Layer Composition Using MLN} \label{sec:modeling-and-computation}
\subsection{Data Modeling}
\label{subsec:modeling}

\begin{wrapfigure}{r}{0.6\textwidth}\centering
 \centering
\vspace{-20pt}	
\includegraphics[width=0.6\textwidth]{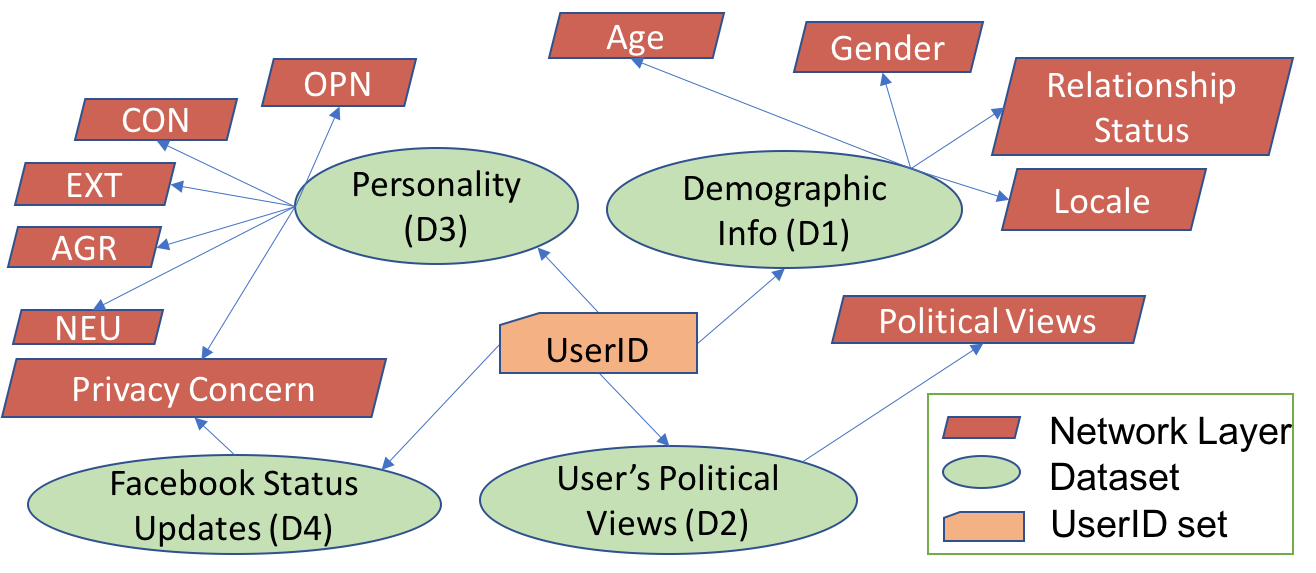}
 \vspace{-10pt}
\caption{Modeling the FB Data Collection for Multilayer Network Analysis}
	\label{fig:data_rdf} 
 \vspace{-30pt}
\end{wrapfigure}

It is already explained in Section \ref{sec:problem-statement} that we have four Facebook datasets (D1-D4) and the corresponding features. \emph{UserID} is common to all datasets to associate with each other as shown in Figure \ref{fig:data_rdf}. \emph{UserID} is also used to make the dataset anonymous to mask the privacy of an individual.




Unlike other features from D1 and D2, the features of personality traits (OPN, CON, EXT, AGR, NEU) and privacy-concern are not explicitly present in the given  dataset and will be derived through content analysis.  The derived output for each personality trait would be binary label ``Yes'' or ``No''.  And the derived output for privacy-concern of each FB user in our datasets is ``high'', ``medium'', or ``low''.  This clearly indicates the power of the approach in absorbing derived content in the same way as an explicit feature. Different types of content extraction can be supported readily. Our approach to content extraction is presented in Section ~\ref{sec:content-analysis}.   

Table \ref{tbl:simple_statistics} shows the statistics regarding the generated multiplex layers along with the number of edges in each layer. The number of nodes in all layers is the same but the number of edges will be different since it depends on the available information for each feature. The 11 layers in Table \ref{tbl:simple_statistics} that are generated for the Facebook multilayer network correspond to the features in Figure~\ref{fig:data_rdf}. The semantics of the graphs in each layer are described as follows:

\begin{table}[h!t]
\centering
\vspace{-18pt}
\caption{Statistics of 11 multiplex layers}
 \scalebox{0.85}{%
 
 		\begin{tabular}{m{3.8cm}|m{1.3cm}|m{1.5cm}}
 			\hline 
 			\textbf{Layer} & From dataset & \# edges \\ \hline
 			L1: Age	&	D1 &	1,228,223 \\
 			L2: Gender	&	D1	& 1,813,638 \\
 			L3: Relationship Status	&	D1 &	1,119,592 \\
 			L4: Political Views	&	D2	& 494,974 \\
 			L5: Locale	&	D1 &	2,799,160 \\
 			L6: OPN	&	D3 &	1,020,306 \\
 			L7: CON	&	D3	& 840,456 \\
 			L8: EXT	&	D3	& 795,691 \\
 			L9: AGR	&	D3	& 718,201 \\
 			L10: NEU	&	D3 &	627,760 \\
 			L11: Privacy Concern	&	D3, D4 &	2,191,659 \\
 			\hline
 		\end{tabular}
}
 	\label{tbl:simple_statistics}
\vspace{-26pt}
\end{table}

\begin{description}
	\item (L1) \textbf{Age}: Any two users are connected by an edge when they both fall into a same age group, namely [$\leq$ 20], [21-30], [31-40], [41-50], [51-60], and [$\geq$ 61].
	\item (L2) \textbf{Gender}: Any two users with the same gender are connected.
	\item (L3) \textbf{Relationship Status}: Any two users with the same relationship status are connected by an edge.
	\item (L4) \textbf{User-Defined Political Views}: Any two users with the same political view are connected by an edge.
	\item (L5) \textbf{Locale}: Any two users with the same locale settings (e.g., en\_UK, en\_US) are connected by an edge.
	\item (L6-L10) \textbf{\PModel (i.e., OPN, CON, EXT, AGR, NEU)}: Each personality trait of the \PModel forms one network layer. Any two users with the same type of personality trait are connected.
	\item (L11) \textbf{Privacy Concern}:  Any two users with the same privacy-concern level (i.e., high, medium, or low) are connected.
\end{description}

%% file: ses4_modeling_and_approach_abh.tex
\subsection{Metric Computation \& Layer Composition}
\label{subsec:computations}

In the multilayer network described above, 
although each layer has the same nodes, their edge connectivity will vary according to the feature value distribution. For example, groups of people having the same political view may not be present in the same age group, groups of people with the same personality may have different levels of privacy-concern and so on. For detecting the tightly connected groups of people with respect to a particular feature, we compute \textit{communities} in the corresponding layer by applying Infomap\cite{InfoMap2014}. A community in a graph translates to a group of nodes that are more connected to each other than to other nodes/communities in the graph.  

\begin{figure*}[ht]
\vspace{-20pt}
	\centering
	\includegraphics[width=0.55\textwidth]{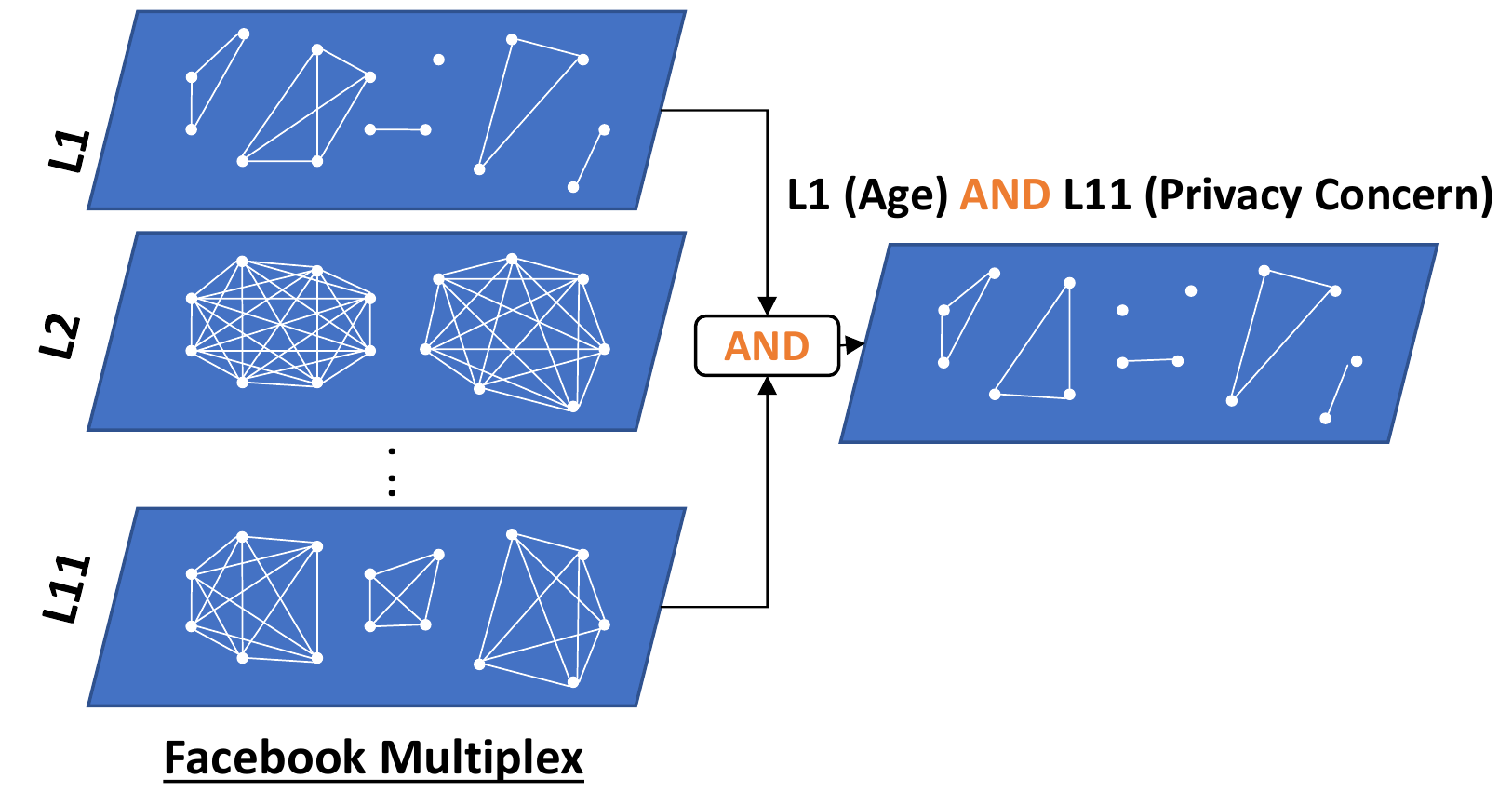}
 \vspace{-10pt}
\caption{An example of AND-Composition}
	\label{fig:ANDComposition}
\vspace{-20pt}
\end{figure*}

Analytical queries listed in Section \ref{subsec:query}, require commonality of information in \textit{at least two multiplex layers}. This corresponds to the Boolean \textit{AND} operation for composing layers. 
For example, in Q4a (Section \ref{subsec:query}), to  analyze the effect of age on level of privacy-concern, we need to compute communities where an edge represents people who fall into the same age group \textbf{and} have the same privacy-concern level. Figure \ref{fig:ANDComposition} shows a simple example with several small Facebook multiplex layers, and an \textit {\textbf{AND}} composition of L1 (Age) and L11 (Privacy Concern) layers. 
Table \ref{tab:computationLayers} shows the \textit{AND}-compositions whose \textit{communities} have to be generated in order to perform the flexible analysis of the Facebook multilayer network for our analytical queries listed in Section \ref{subsec:query}. 

\begin{table}
\centering
	    \caption{\textit{AND}-Compositions needed for analytical queries shown in Section \ref{subsec:query}}
		\scalebox{0.9}{
        \begin{tabular}{|m{0.8cm}|m{7.2cm}|}
            \hline
            \textbf{Ana-lysis} & \textbf{Required \textit{AND}-Compositions of Layers} \\
            \hline
            \hline
            \multicolumn{2}{|l|}{\textit{Dominant Political Views}} \\
            \hline
 \textbf{Q1} & L1 (Age), L4 (Political View), L5 (Locale)\\
            \hline
            \hline
            \multicolumn{2}{|l|}{\textit{Relationship Status Correlation}} \\
            \hline
\textbf{Q2a} & L1 (Age), L3 (Relationship Status), L5 (Locale)\\
            \hline
\textbf{Q2b} & Five 3-layer compositions: [L2 (Gender) AND L3 (Relationship Status)] with each of L6 (OPN), L7 (CON), L8 (EXT), L9 (AGR), L10 (NEU)\\
            \hline
            \hline
            
            \multicolumn{2}{|l|}{\textit{Personality Traits Analysis}} \\
			\hline
\textbf{Q3a} & L6 (OPN), L10 (NEU)\\
            \hline
\textbf{Q3b} & Five 2-layer compositions: Each of L6 (OPN), L7 (CON), L8 (EXT), L9 (AGR), L10 (NEU) with L1 (Age)\\
            \hline
            \hline
            \multicolumn{2}{|l|}{\textit{Privacy Concern Correlation}} \\
			\hline

\textbf{Q4a} & L1 (Age), L11 (Privacy Concern)\\
            \hline
\textbf{Q4b} & Five 3-layer compositions: [L2 (Gender) AND L11 (Privacy Concern)] with each of L6 (OPN), L7 (CON), L8 (EXT), L9 (AGR), L10 (NEU)\\

            \hline
\end{tabular}
}
\label{tab:computationLayers}
\vspace{-10pt}
\end{table}

\noindent\textbf{Community Detection for \textit{AND} Compositions:} The traditional way to address any of the listed analytical queries is to first generate the required combined graph corresponding to the \textit{AND}-composition and then detect the communities. 
The MLN approach pre-computes the 
communities of the individual layers. Based on the composition requirements of the analytical query, partial results (i.e., pre-computed communities) are intersected (for \textit{AND}-composition) to generate the combined communities. It can be shown analytically  ~\cite{ICCS/SantraBC17} that both computationally and storage-wise, the MLN approach is more efficient.
The requirement to apply composition is that the communities of the individual layers must be \textit{self-preserving} in nature.


\noindent\textbf{Checking the Self-Preserving Property:} By definition, \textit{``a community is self-preserving if the nodes in it are so tightly connected such that even if only a subset of connected nodes are chosen from that community, they will form a smaller community~\cite{ICCS/SantraBC17}."} For each of the 11 layers, we computed the internal clustering coefficients of each node, which is the ratio of the number of edges among the neighbors of the node \textit{in the same community} to the total possible edges among those neighbors, and it was \textit{one}. This indicated that the \textit{communities in each layer were self-preserving}. 
In Section \ref{sec:analysis-results}, we will discuss some of the inferences that have been drawn from this analysis, and in Section \ref{sec:efficiency-evaluation} we highlight the performance analysis between the traditional and MLN approaches.

%% file: ses5_content_analysis_svx.tex
\section{Content Analysis on User Generated Content} \label{sec:content-analysis}

As briefly introduced in Section \ref{sec:modeling-and-computation}, we apply content analysis to derive the features of five personality traits and privacy-concern to adapt into the proposed multilayer network (MLN) approach. For detecting personality traits, ~\cite{Pennebaker:99} has identified many linguistic features associated with each personality trait in \PModel. For instance, \emph{Extraversion} {\small(EXT)} tends to seek stimulation in the external world, the company of others, and to express positive emotions. \emph{Neuroticism} {\small(NEU)} people use more 1st person singular pronouns, more negative emotion words than positive emotion words. As mentioned earlier in Section \ref{subsec:modeling}, five personality trait scores for each user are given in the dataset D3. To form layer L6-L10 using these personality traits in the proposed multilayer network analysis,  we followed the strategy from \cite{Celli:2013} by using five mean-values \{3.8, 3.5, 3.6, 3.55, 2.8\} of the given personality scores to decide ``Yes" or ``No" label of \{OPN, CON, EXT, AGR, NEU\} correspondingly. ``Yes'' label is assigned if the personality trait score is higher than or equal to the corresponding mean-value. Otherwise, ``No'' label is assigned. In the following of this section, we will mainly explain how to generate privacy-concern as one layer to adapt the multilayer network.  

Previous work~\cite{Vu:2018b} has found that using given personality and status updates of users, privacy-concern can be predicted accurately based on UGC data.  We extended their model by building a deep neural network model to automatically classify users' privacy-concern to high (\HighConcern), medium (\MediumConcern), and low (\LowConcern) based on given five personality trait scores and status updates. Motivated from previous studies~\cite{Pennebaker:99,Vu:2018b,ThanhVu:2018}, we extract the following content features from Facebook status updates, which will be used in the following deep neural network model to predict users' privacy-concern:

\begin{itemize}[wide]
	\item \textbf{Polarity features:} Since sentiment words embody personality (e.g., \emph{``I do\textbf{n't} \textbf{hate} you. Well, you found me $\cdots$"}), we use the number of polarity signals appearing in FB status updates as the polarity features. We identify positive and negative words using the sentiment dictionaries provided by Hu and Liu \cite{hu2004mining}. Additionally, we consider boolean features to check whether or not a negation word is in a FB status (e.g., n't).
	\item \textbf{Syntactic features:} We extract part-of-speech tags (POS tags) for all status updates in the dataset. Afterwards, we use all the POS tags with their corresponding term-frequency and inverse-document-frequency (tf-idf) values as our syntactic features and feature values, respectively.
	\item \textbf{Semantic features:} The major challenges when dealing with user generated data are: (1) the lexicon used in a status update is informal with many out-of-vocabulary words and (2) they are usually short text \cite{XuanSon:2018}. The lexical and syntactic features seem not to capture that property very well. To handle this challenge, we apply two approaches to compute
vector representations for FB status updates. First, we utilize Latent Dirichlet Allocation (LDA) \cite{Blei:2003} for discovering the abstract ``topics" that occur in all FB status updates. Secondly, we employ 300-dimensional pre-trained embedding models at word level \cite{Mikolov:2013} and at character level \cite{Kim:2016} to compute a representation for a FB status update as the average of the embeddings of words and characters in the status update.
	\item \textbf{Lexical features:} include [\textit{1-5}]-grams in both word and character levels.
For each type of \textit{n}-gram, we only select the
top 1,000 \textit{n}-grams based on \textit{tf-idf}.
\end{itemize}

\begin{figure}
	\centering
\includegraphics[width=0.6\textwidth]{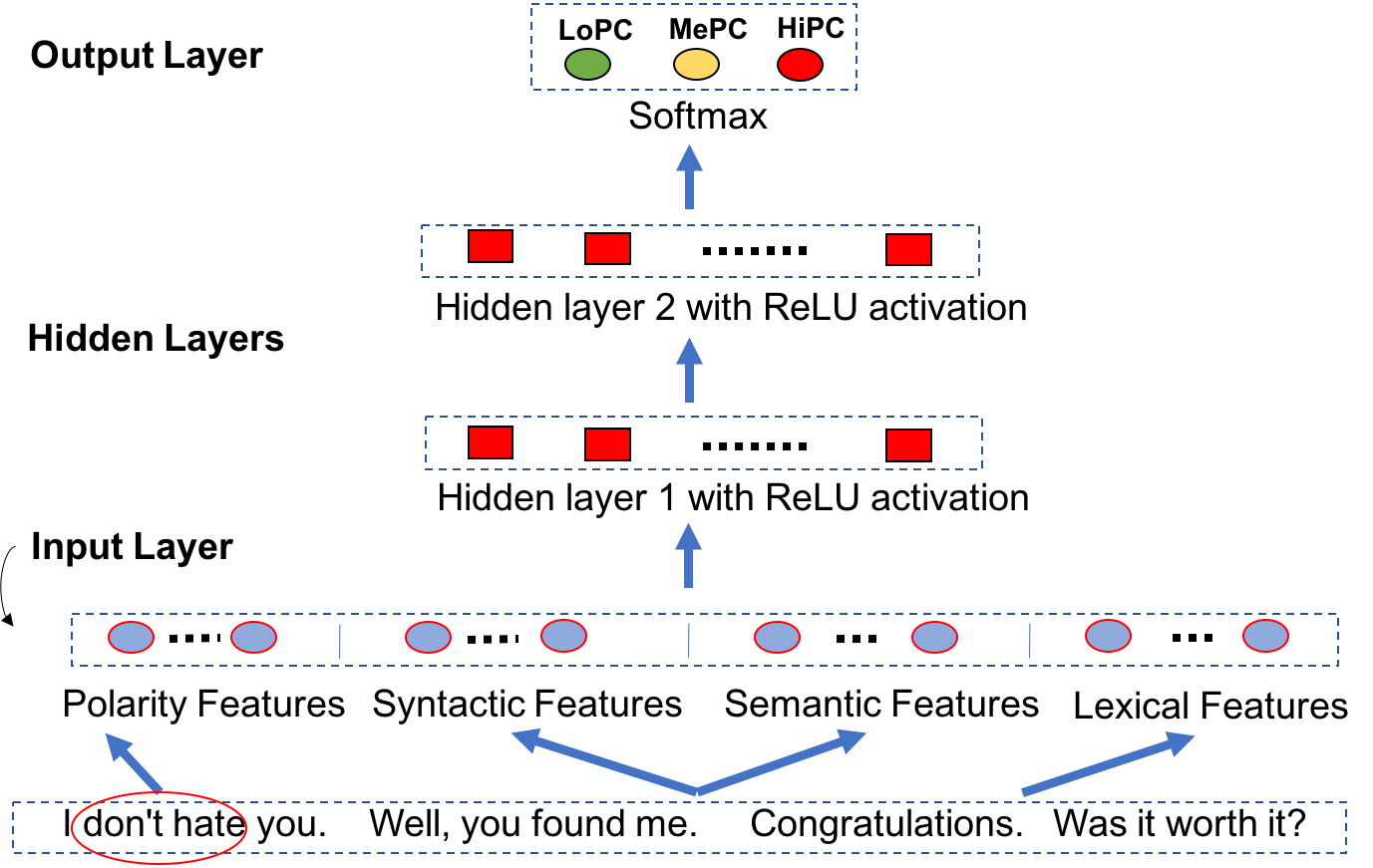}
	\vspace{-0.4cm}
\caption{Neural network model for privacy-concern detection}
	\label{fig:dnn_architecture}
\vspace{-18pt}
\end{figure}

\textbf{Deep Neural Network Model}: The recent novel model from \cite{Vu:2018b} was adapted to find users' privacy-concern level based on their social network status updates or personality-trait score.  Their neural network model was proposed mainly for privacy-degree prediction, instead of privacy-concern level prediction. To adapt the proposed multilayer network, there is a need of categorized output to create a privacy-concern network layer (L11), where users with the same privacy-concern level is supposed to be connected.  Thus, we extended their approach by developing a category-based privacy-concern detection model as shown in Figure \ref{fig:dnn_architecture}.  This extension does not only matter to the output (from discrete to categorization), but the feature representations in hidden layer of deep neural network model are different.


It is a Multilayer Perceptron
(MLP) model \cite{Hornik1989}, the architecture of which consists of an input layer, two hidden layers and a softmax output layer.  Given all Facebook status updates of a user, the input layer represents the status update by a feature vector which concatenates lexical, syntactic, semantic and polarity feature representations. The two hidden layers with ReLU activation function \cite{Nair:2010} take the input feature vector to select the most important
features which are then fed into the softmax output layer for privacy-concern level detection and classification. Regarding classification performance evaluation, we split 20\% of the data for a blind test. We run 10 fold cross-validation on the rest 80\% to train and select the best hyper-parameters. After all, the model achieved an accuracy of 84.44\% on the blind test set. Furthermore, we compare the model with other popular models (i.e., support vector machine, random forest) and the recent advanced model (i.e., C-LSTM \cite{ZhouSLL15b}), and none of them performs as good performance as our model does. This clearly shows the effectiveness of the model to predict privacy-concern levels based on UGC data. As explained above, the predicted user privacy-concern levels are used to create the network layer L11 of the multilayer network. 


%% file: ses6_analysis.tex
\section{Experimental Study and Query Results Analysis}
\label{sec:analysis-results}
\input{ses6_analysis_results_abh}


\section {Efficiency Analysis of The MLN Approach}
\label{sec:efficiency-evaluation}
\input{sc_sec7_efficiency_new}

%% file: ses6_analysis_results_abh.tex
This section shows the results for each analytical queries (Q1-Q4) formulated in Section \ref{subsec:query}. Based on the communities obtained for the required \textit{AND}-Compositions listed in Table \ref{tab:computationLayers}, a detailed analysis was performed to draw some insights that are discussed below. 
To the extent possible, we have related our analysis with various published independent surveys.

\begin{figure}
\vspace{-30pt}
\centering
\begin{minipage}{.48\textwidth}
  \centering
  \includegraphics[width=\textwidth]{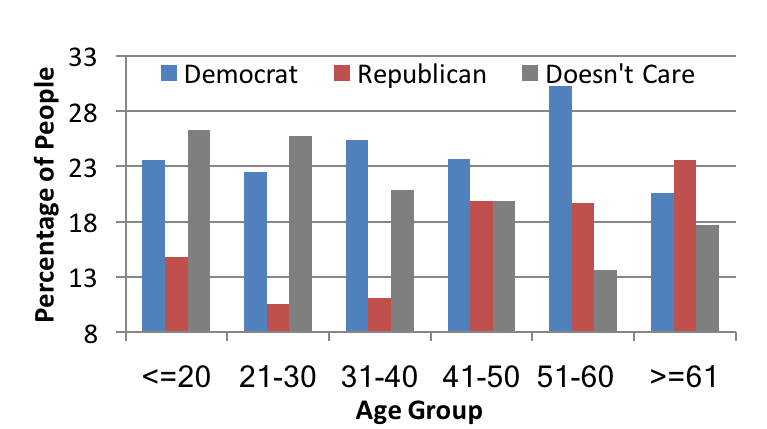}
  \captionof{figure}{Top 3 political views by age group}
  \label{fig:Q1a}
\end{minipage}%
\hspace{0.1cm}
\begin{minipage}{.48\textwidth}
  \centering
  \includegraphics[width=0.85\textwidth]{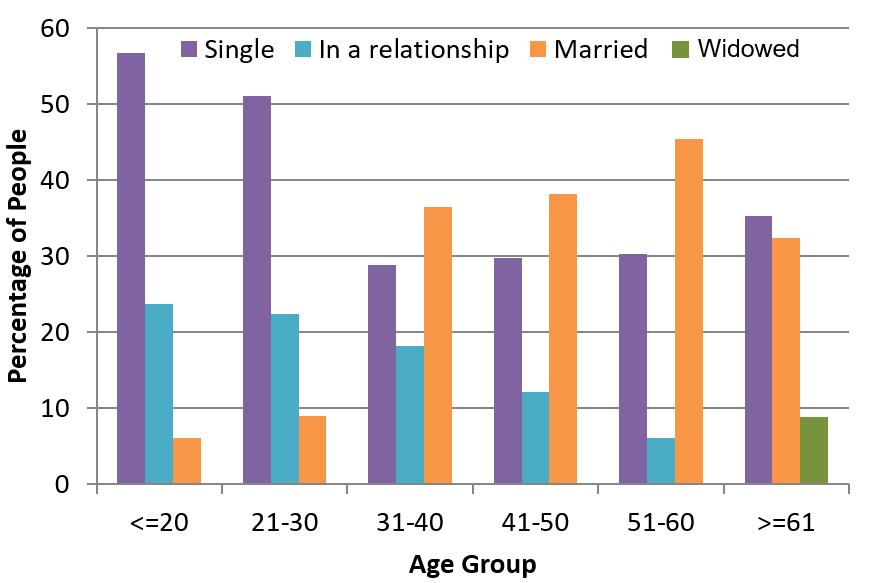}
  \captionof{figure}{Top 3 relationship statuses by age group}
  \label{fig:Q2a}
\end{minipage}
\vspace{-20pt}
\end{figure}

\textbf{Dominant Political Views (Q1):} Figure \ref{fig:Q1a} shows the distribution of the top three political views over the US population active on Facebook for each age group. Some observations are:

\begin{itemize}[wide]
\item Among the politically interested and socially active US people across age groups, the majority supported the democrats in the period of 2007-2012. As we know, there was a lot of support for the democratic presidential candidate who got elected in 2008, and we believe this is reflected in the political views of that period. The period of 2007-2009 also indicated the same which makes sense as the campaign was underway in that period. This is confirmed in \cite{Q1Elections1,Q1Elections2} since Barack Obama, a democrat, who took the US presidency on January 8, 2009 was able to influence people's political leanings through his presidential campaign from February 10, 2007. 

\item Among the socially active youth ($\leq$ 30 years old), majority of them have the political view of ``doesn't care". Although this includes people below the voting age, even the published statistics \cite{Q1Elections3} show that young people are least likely to vote and may not have formed an opinion about their political leanings.
\item Among politically interested youth ($\leq$ 30 years old) who are also active on social media, dominant support is for democrats. This may be attributed to a few of president Obama's youth centric movements and the significant use of social media for the first time in a US election and also to his subsequent accomplishments~\cite{Q1Elections4}.

\item Interestingly, among all age groups, only the ones above 61 years old favored republicans over democrats, which is also reflected in the election reports from 2008 \cite{Q1Elections5}.

\end{itemize}


\textbf{Relationship Status Correlation (Q2):} The preference of a relationship status based on age and the corresponding effect on personalities of different genders were analyzed.
\begin{enumerate}[label=\textbf{(Q2\alph*)},wide]
\item \textit{Variation with Age:}
Figure \ref{fig:Q2a} shows, for each age group, the distribution of people among the group's top three relationship statuses. A few intuitive inferences that can be drawn are:

\begin{itemize}
\item The youth ($\leq$ 30 years old) stay single than be in a relationship or get married, according to the given dataset.
\item The percentage of married people steadily increases with age which can be attributed to the popular fact that as people age, they want to be in a longer term commitment (in a relationship or married).
\item The transitions from ``Single'' to ``In a relationship'' to ``Married'' are clearly seen with change in age in Figure \ref{fig:Q2a} which matches the social trend.
\item The third largest fraction of people in age group ($\geqslant$ 61) constitutes those who have lost their spouses (Widowed).
\end{itemize}

\item \textit{Effect on Personality and Gender:} Changes in relationship status seem to have effects on the personality. Moreover, this change seems to be correlated with the gender. 
For the given population, we present the distribution of males and females among the top three relationship statuses - Single (S), In a relationship (R), Married (M), who display different personality traits in Table \ref{tab:Q2b}. The ones marked in \textbf{bold$^\star$} and \textit{italics$^\dag$} represent the category of people with highest and lowest percentages, respectively. 


\end{enumerate}

\begin{table}
		\centering
	    \caption{\% of people with different personality traits based on relationship status and gender}
        \begin{tabular}{m{1cm}|m{0.85cm}|m{0.95cm}|m{0.95cm}|m{0.85cm}|m{0.95cm}|m{0.95cm}}
            \hline
            \multirow{2}{*}{\textbf{Trait}} & \multicolumn{3}{c|}{\textbf{Males}} & \multicolumn{3}{c}{\textbf{Females}}\\
            \cline{2-7}
            & \textbf{S(\%)} & \textbf{R(\%)} & \textbf{M(\%)} & \textbf{S(\%)} & \textbf{R(\%)} & \textbf{M(\%)}\\ \hline
OPN & \textbf{55.7$^\star$} & 54.8 & 47.3 & 53.0 & 55.6 & \textit{43.0$^\dag$} \\
CON & \textit{44.2$^\dag$} & 50.2 & 52.7 & 46.1 & 51.7 & \textbf{58.5$^\star$} \\
EXT & 45.0 & \textbf{54.8$^\star$} & \textit{35.5$^\dag$} & 49.1 & 50.8 & 44.6 \\
AGR & 42.9 & 50.2 & \textit{31.8$^\dag$} & 46.4 & 45.9 & \textbf{50.4$^\star$} \\
NEU & 36.5 & \textit{26.6$^\dag$} & 33.6 & 45.2 & \textbf{53.8$^\star$} & 48.1 \\
            \hline            
\end{tabular}
\label{tab:Q2b}
\end{table}

A few observations that can be made from the above analysis are: a) Married females have least openness (OPN) to experience, and highest conscientiousness (CON). b) Married females have the highest agreeableness (AGR), while married males have the least.  These observations have been made with respect to the given population and need to be confirmed on a bigger population.

\textbf{Personality Traits Analysis (Q3):} 
Various analyses based on the five personality traits for each individual are discussed below.

\begin{enumerate}[label=\textbf{(Q3\alph*)},wide]

\item \textit{Contrasting personality traits:} Clearly, if a person feels anxious and does not stay relaxed (NEU) then he/she will try to make his/her life comfortable by indulging in less stressful activities making them be less open to new experiences (OPN). Thus, the fraction of people displaying these contrasting personality traits is supposed to be low. Our analytic results go hand in hand with this intuition as just 23.3\%  people belonging to this category. 

\item \textit{Personality trait evolution with Age:}  Figure \ref{fig:Q4b} shows how each personality trait varies with age, based on which, few interesting observations can be made as follows.



\begin{figure}
\centering
\begin{minipage}{.48\textwidth}
\centering
  \includegraphics[width=0.9\textwidth]{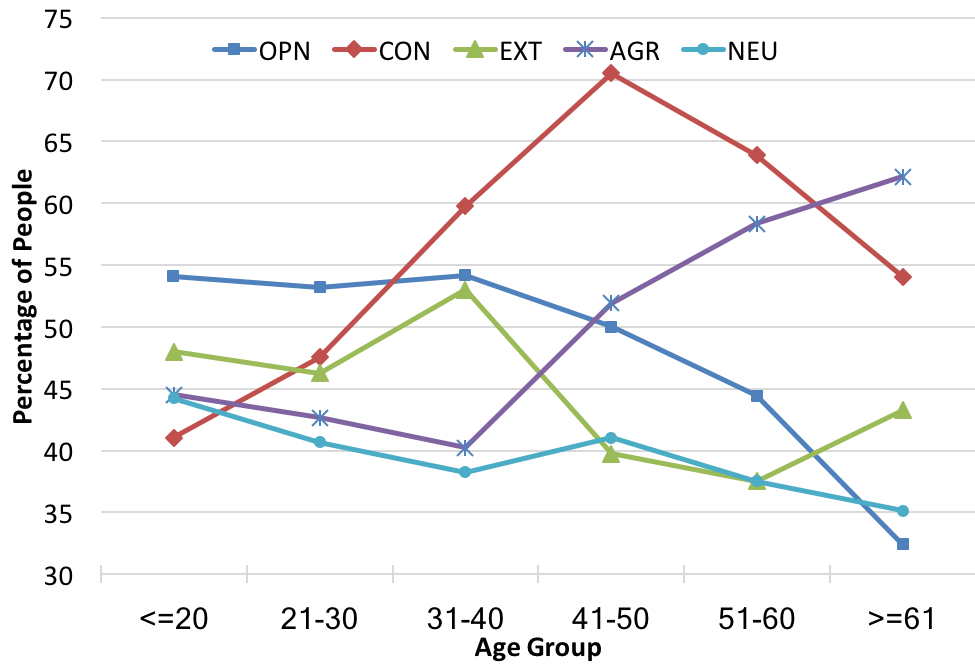}
  \captionof{figure}{Changing personality distribution with age}
\label{fig:Q4b}
\end{minipage}%
\hspace{0.1cm}
\begin{minipage}{.48\textwidth}
  \centering
  \includegraphics[height=0.18\textheight,width=0.31\textheight]{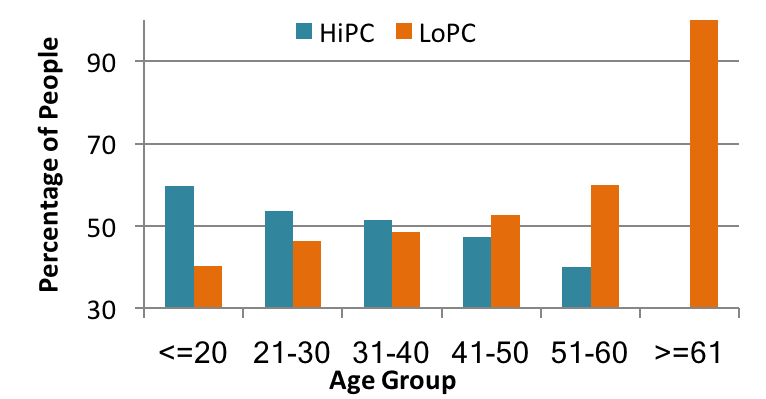}
  \captionof{figure}{Distribution of HiPC and LoPC by age group}
  \label{fig:Q3a}
\end{minipage}
\vspace{-20pt}
\end{figure}


\begin{itemize}[wide]
\item  Openness (OPN) reflects whether one prefers new experiences and to engage in self-examination. This trait increases with age and peaks around the 30s (54.2\% in age group of 31-40). However, older people prefer to go with the tried-and-tested approach (67.6\% of the people above 60 years old resist new experiences). 

\item Conscientiousness (CON) associates with achievement and working systematically, methodically and purposefully. Analysis shows that the age group with conscientiousness the most is 41-50 years old. A recent survey about founders and entrepreneurs indicated that their average age was 45 years  old~\cite{azoulay2018age}.  
 
\item Extraversion (EXT) describes one's sociability and enjoy to be the center of attention. This trait seems to peak at two age groups (i.e., [31,40] and [$\geqslant$ 61])  in the dataset. 
 
\item Agreeableness (AGR) reflects a tendency to perceive others in a more positive light. Parenthood and grand-parenthood may make the elder generation more empathetic towards others as compared to the younger lot.

\item Neuroticism (NEU) reflects one's ability to deal with emotion states, such as stress and anxiety.  It can be observed from Figure \ref{fig:Q4b} that the younger lot does not deal very well with stress. Even the studies substantiate this finding as around 80-90\% adolescent suicides are linked to common psychiatric disorders, such as depression and anxiety \cite{cash2009epidemiology}. This trait (NEU) seems to be most stable over age compared to other traits.
\end{itemize}

\end{enumerate}

\textbf{Privacy Concern Correlation (Q4):} Three levels of privacy concern (PC) have been considered for this work - \HighConcern, \MediumConcern and \LowConcern. Here we have taken into account age, gender, and personalities as three parameters for performing analysis to understand the choice of particular level of privacy-concern.
\begin{enumerate}[label=(\textbf{Q4\alph*)},wide]
\item \textit{Variation of privacy-concern across age groups:} The concern level of sharing personal information on the social media varies with age. Irrespective of the age group, the MePC was the most dominant level of privacy. Out of the remaining individuals, Figure \ref{fig:Q3a} shows the distribution of the people with extreme levels of privacy - High and Low. Few observations are discussed below.

\begin{itemize}[wide]
\item People ($\leq$ 40 years old) prefer the higher level of privacy. This can be attributed to the fact that this age group is probably more aware of the cons of sharing sensitive personal information on the web such as identity theft. 
\item The status updates of people ($\geqslant$ 41 years old) contain more personal information and this trend increases with age. This reflects a lower level of privacy-concern probably due to their unawareness of the potential harm from disseminating personal information on social media. 
\end{itemize}

\item \textit{Correlation of privacy-concern over personality traits:}   
Table \ref{tab:Q3b} shows the two extreme personality traits and their corresponding privacy-concerns for males and females. 

\begin{table}[h!t]
\centering
	\caption{Dominant personality traits of male and females preferring different levels of privacy-concern}
	\vspace{-0.05cm}
	\scalebox{0.9}{
		\begin{tabular}{m{1.6cm}|m{1.8cm}|m{1.9cm}|m{1.8cm}|m{1.9cm}}
			\hline
			\multirow{2}{*}{\textbf{Privacy}} & \multicolumn{2}{c|}{\textbf{Extraversion(100\%)}} & \multicolumn{2}{c}{\textbf{Neuroticism(100\%)}}\\
			\cline{2-5}
			& \textbf{Males(\%)} & \textbf{Females(\%)} & \textbf{Males(\%)} & \textbf{Females(\%)}\\ \hline
			HiPC &  0 & 0 & 09.90 & \textbf{09.99} \\
			MePC & 39.89 & \textbf{45.36} & 30.15 & \textbf{49.96} \\
			LoPC & 07.45 & 07.30 & 0 & 0 \\
			\hline            
		\end{tabular}
	}
	\label{tab:Q3b}
\vspace{-20pt}
\end{table}

\begin{itemize}[wide]
\item Females have higher privacy-concern than males on both extraversion and neuroticism. This observation kept consistency with the previous study in~\cite{rowan2014observed}.
\item Both males and females on extraversion display low and some medium levels of privacy-concern. This matches the definition of extraversion from social scientists. That is, people who enjoy being the center of attention are likely to share more personal information on the web such as check-ins and day-to-day activity updates. 
\item Both males and females on neuroticism tend to have predominantly high privacy-concern, without anyone having low level privacy-concern. This matches the social scientists' definition. 

Analysis Q4a and Q4b are important as it validates our content extraction approach to derive accurate privacy-concern.

\end{itemize}

\end{enumerate} 


%% file: sc_sec7_efficiency_new.tex
So far, we have established the modeling and flexibility of analysis of the MLN approach. Below, we will establish its efficiency in general and highlight it with respect to the current dataset analysis.

\noindent\textbf{Processing Layers Instead of a Single (large) Graph (SLG): }  Separation into layers allows one to process each layer \textit{only once} for \textit{all analysis} and the composition allows us to make use of the pre-computed partial results. Furthermore, in many cases, the size of each layer is likely to be smaller than the size of combinations of layers. On the other hand, a new combined graph needs to be created and processed in the traditional approach for each \textit{unique} analytical query. Storing each layer is more compact and uses less memory than storing all the required layer combinations.

\noindent\textbf{Processing Layers in Parallel: } The MLN approach easily lends itself to process all (or subsets of) layers in parallel to further improve efficiency. The total cost is the processing cost of the most complex layer. This can only be done for a set of known analysis queries in the traditional approach in contrast to the MLN approach where it needs to be done only once. In the MLN approach, it is further possible to process each layer in parallel by partitioning and leveraging existing algorithms (again one time). Although this can be done in the traditional approach, it has to be done \textit{after} the combined graph is created which reduces its effect significantly.

\noindent\textbf{Efficiency of Composition: } The core of the MLN approach is its ability to compose layers pair-wise to get complete, correct results. Each composition is likely to be on fewer and smaller number of components (from each layer) thereby reducing the resources needed (both storage and processing).

\noindent\textbf{Combinatorial Reduction: } As the complexity of dataset increases, it translates to more layers (corresponding to more features). Based on the number of layers, in the MLN approach, there is a significant and non-linear reduction in the processing cost as the number of layers increase as compared to the traditional approach. Assuming $M$ layers, for an exhaustive analysis, they need to be combined in $2^{M}-1$ ways, each representing a unique analysis of a combination of features. This translates to creating $2^{M}-1$ combined graphs in the traditional approach and processing them individually. In the MLN approach, instead, $M$ layers are processed \textit{once} and $2^{M}-1$ combining of partial results from layers are performed. As we show below, these compositions are significantly smaller (by orders of magnitude) computationally as compared to processing of a layer. Essentially, the exponential complexity has been reduced to a linear one one with very little additional processing.

\subsection{Complexity Analysis}
\label{subsec:complexity}

For the complexity analysis, we assume that for a given multilayer network with fixed number of $M$ layers, say $\{G_1, \cdots, G_M\}$, each of the $N$ query analyses requiring ${K}$ related layers on average, should return a list of communities, $L$. 
\begin{itemize}[wide]
\item Single  (large) Graph (SLG):  In this approach, for every analysis it first generates the composed graph, $G_{AND}$, obtained through ($K$-1) 2-layer \textit{AND}-compositions, on average. On this \textit{AND}-composed layer, we apply the Infomap technique (InfoM) \cite{InfoMap2014} to generate the list of communities, $L$. Thus, for $N$ analyses the complexity of this approach will be $O(N*(AND_{i=1}^{K} G_i  + \text{InfoM}(G_{AND}))$, where $K \leq M$. 
\item MLN: Its first step is to generate the communities for each of the $M$ layers by applying Infomap. While generating the communities we also obtain the internal clustering coefficient for each node, which are used to determine if the communities are \textit{self-preserving} or not. 
If the property of \textit{self-preserving} is satisfied, then for each analysis, to generate the list of communities $L$, for the corresponding \textit{AND}-composition $G_{AND}$, communities from \textit{K} layers are intersected based on nodes. Thus, for $N$ analyses the cost of this approach will be, $O (\sum_{i=1}^{M} (\text{InfoM}(G_i)) + N*\cap_{j=l}^{K-l+1} C_{j})$, where $K \leq M$. 
\end{itemize}

In terms of storage space, one needs to store $M$ lists of communities for {MLN}, whereas in {SLG}, $N*(AND_{i=1}^{K} G_i)$ graphs need to be stored. Considering both space and time, if the number of analyses, $N$, and the average number of layers required for each analysis, $K$, are low then the one-time cost of performing $M$ number of Infomap operations in {MLN} will dominate and make {MLN} more expensive as compared to {SLG}. However, with the increasing values of $N$ and $K$, the efficiency of {MLN} over {SLG} improves significantly, as the cost of producing the number of {AND}-composed graphs and applying Infomap that traverses through the edges of each of them, begins dominating. In the worst-case for $N$ = $O(2^M)$ and $K$ = $O(M)$, {SLG} will perform an exponential number of \textit{AND}-Compositions and edge traversal based Infomap operations whereas, {MLN} will just perform the cost-effective node intersection of layer-wise communities.


\subsection{Computational Results}
Below, we show efficiency results of analysis on the given Facebook datasets.

\noindent\textbf{Experimental Set up:} We used a quad-core 8th generation Intel i7 processor Linux machine with 8 GB memory for all of our analysis. Based on Table \ref{tab:computationLayers}, we computed communities for a total of 19 \textit{AND}-compositions to answer the queries Q1 to Q4, each requiring 3 layers on an average.


\begin{figure}
\vspace{-15pt}
\centering
\includegraphics[width=0.71\textwidth]{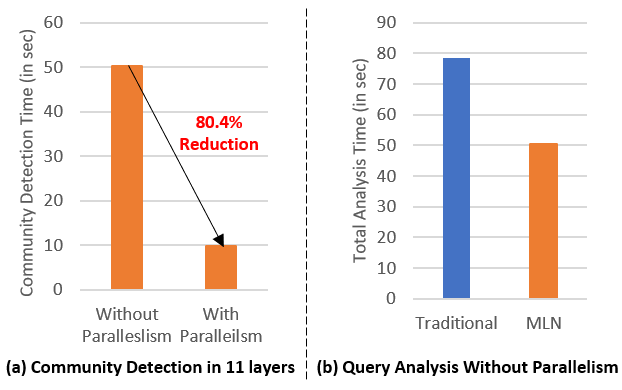}
\caption{Efficiency Comparison of MLN and Traditional Approaches}
	\label{fig:eff}
\vspace{-20pt}
\end{figure}

Figure~\ref{fig:eff} (a) shows the time taken for processing all of 11 layers \textit{with and without parallelism}. As can be seen, in the MLN approach \textit{with parallelism}, it reduces the cost of processing the most complex layer (9.847 seconds for L5: Locale, 2.8 million edges, Density: 0.77 - \textit{most dense}) -- a \textit{\textbf{reduction of 80.4\%} approximately}. 

The incremental computation cost for each query using the MLN approach is extremely small. This can be appreciated from the worst case scenario - comparing minimum layer processing cost with maximum composition cost. The total  composition cost to answer the most complex query (Q2b) was \textbf{0.039 seconds} and the minimum layer processing cost was \textbf{1.61 seconds} (L4: Political View, 494K edges, Density: 0.14 - \textit{least dense}). \textbf{The difference is more than two orders of magnitude.}


Figure \ref{fig:eff} (b) shows the \textit{total time} taken to answer the analysis queries using the traditional and the MLN approach, respectively, \textit{without parallelism}, as 78.520 seconds and 50.354 seconds (for \textbf{36\% reduction}). Further, if communities for each layer are generated in parallel, total computation time for the MLN approach reduces to 9.987 seconds (for \textbf{87\% Reduction}). Also, note that the analysis shown in this paper is \textbf{less than 1\% of the possible analysis}.

In summary, the experiments on the Facebook Dataset validate the \textit{MLN approach from an efficiency perspective as compared to the traditional approach}.

%% file: ses7_conclusions.tex
\section{Conclusions}
\label{sec:conclusions}
In this paper, we have applied the emerging MLN approach model and analyze a social network data collection in a flexible and efficient way. We have also shown how content analysis can be readily incorporated into the proposed MLN approach. Experimental analysis and evaluation not only demonstrate the flexibility and efficiency of data analysis using the MLN approach but also validate the analysis results. For future study, we plan to (1) apply the proposed MLN approach to the bigger full data collection and (2) apply hypothesis testing on two different data distributions (e.g., the current one versus the full data collection) to see the statistical significant degree of our findings.

\section*{Acknowledgement}
This work is supported by the STINT project funded by The Swedish Foundation for International Cooperation in Research and Higher Education. The authors also thank the myPersonality project for data contribution.